# Utilizing distilBert transformer model for sentiment classification of COVID-19's Persian open-text responses

**Fatemeh Sadat Masoumi[1], Mohammad Bahrani[2]**

1- Department of Computer science, Allameh Tabatabe'I University
2- Department of Computer science, Allameh Tabatabe'I University
Fatemeh_masoumi@atu.ac.ir

**Abstract**
The COVID-19 pandemic has caused drastic alternations in human's life in all aspects. The government's laws in this regard affected the lifestyle of all people. Due to this fact studying about the sentiment of individuals is important to be aware of the future impacts of the coming pandemics. To contribute to this aim, we proposed a NLP (Natural Language Processing) model to analyze open-text answers in a survey in Persian and detect positive and negative feelings of the people in Iran. In this study, a distilBert transformer model was applied to take on this task. We deployed three approaches to perform comparison, and our best model could gain accuracy: 0.824, Precision: 0.824, Recall: 0.798 and F1score: 0.804.

## 1. Introduction

The very first cases of Coronavirus disease 2019 (COVID 19) were reported in December 2019 in China (Chahrus et al. 2020) [1]. In order to decrease the number of both short and long traveling, the Chinese government introduced some lockdown policies. A short time after that other countries reported the presence of infected patients with this contagious disease in their countries, and World Health Organization (WHO 2020) announced COVID19 as a pandemic global issue [2]. Public authorities with the objective of having control on this pandemic introduced different obligations ranging from travel bans and social distancing to banning social gatherings and closing restaurants and shops. Such this conditions are correlated with physical and mental health of the society.

During this interval, self-reported depression and anxiety were on the increase. It has been shown in the study by Yen et al. that 13% beginning or increasing their use of various substances to help them coping with the situation, 31% reported depression and stress, and 26% reported for having stress related disorders (CDC 2020; Yan et al. 2020) [3][4]. Training centers also have been impacted and were closed, which led into social isolation among the young and the adult. Based on the introduced restrictions they did not have any entertainment to relief their anxiety and stress.

The opportunity to deploy NLP has opened a new window of applications. A plethora of data can be analyzed sentimentally in short time by the artificial intelligence models, with the aim of making decisions. In this study, we have proposed our NLP approach to work with a large number of open-text answers in a survey, instead of using human forces for analyzing them, which is a more expensive approach in terms of money and time. The same work was done by Jojoa et al. [5], but it does not contain the Persian responses' classification, so to the best of our knowledge we are the first group who did it for Persian open-text answers to COVID19 questionnaire.

This paper contains six sections. In section 1, a theme of the effect of COVID19 on the day-to-day lifestyle of the Iranian students is introduced. Section 2 introduces the methods utilized to develop the work, distilBERT transformer has been deployed as the classifier model and it was pretrained by GLUE (General Language Understanding Evaluation Benchmark) database and fine-tuning with our data. In section 3, the steps of the used method are described, and in Sec. 4 obtained results are presented, which are discussed later in section 5, and the conclusion is drawn in Sect. 6.

## 2. Methods and Materials

An invitation was sent to different faculties and students' associations to take part and support data collecting throughput filling out a questionnaire. This study include volunteers affiliated with a post-secondary intuition for example students, staff and faculty. The Survey was available for three months.

### 2-1. Data Description

The data was gather from students and staff and faculties throughout their response to some questions. Each dataset compromises of two fundamnetal columns, the textual response and its related sentiment class. This response was the participant's answer to some queries about how quarantine during COVID19 impact their mental health.

The data used for this study gathered from 11 high schools and 3 universities. There were three various types of questions in the questionnaire: quantitative, qualitative, and open-text ones (Nowrouzi-Kia et al. 2022) [6]. In order to analyze the feeling pf individuals who took part in completing this questionnaire for open-text questions we used NLP methods.

One the very important task in Supervised Machine Learning paradigms is labelling. The labels were obtained from a group of experts who judged the feeling included in each text. To alleviate the risk of biased labelling we did it in a committee of 4 psychologists. And we classified the feeling in each text into two categories: positive and negative.

## 3. Methods

To take on the work, the 4 stages are taken: data pre-processing, data labeled by experts, fine-tuning distilBERT, and metrics measurement.

### 3-1. Data Pre-Processing

The quality of data employed to train the model could be a game changer element in regard to the model's performance. Thus before using the data for fine-tuning it should be cleaned and null and blank values should be eliminated. Some participants in questionnaire had not answered the open-text questions so the null value was shown in our Data Frame, eliminating null values guarantees no stoppage during fine-tuning of the machine learning model. Furthermore, all text was changed to capital letters so as to omit the case sensitivity problem of the model. Also, answers with unclear ideas were omitted as well.

### 3-1-1. Unifying Language of Dataset

All data was translated to English, the idea behind this was to generalize the textual data to a language that can be understood and fine-tuned by a greater number of people. Also, having a unified model can keep consistency throughout the dataset and having an enormous English dataset with textual record for training which is classified into its related sentiment to use English as a language and even concatenate our translated dataset to other related English dataset and make the cumulative dataset bigger. Besides, finding valid sources to train model for Persian could be proved difficult.

For translating, Google API was used, by each API textual responses in multiple batches can be translated. One instruction to Google translate application is the API call to interpret our text to code for the learning model, since the Google's Cloud services is using throughout this interpration process there is not any requirements for high-tech hardware.

### 3-2. Data Labelling by Professionals

The transformer based method is a supervised method (Karpov et al. 2019) (Conneau et al. 2017)[7] [8], so every input text is required to have a label. A group of four human professionals were employed to judge the sentiment of each text. The process was simple, every expert read a text and based on his knowledge judge whether it involve positive feeling or negative one, at the end when they reach a verdict about the class of that text they will label the textual answer.

### 3-3. Fine-Tuning Distilbert for Sentiment Classification

One of the machine learning framework for Natural Language Processing is Bidirectional Encoder Representations (BERT) (Lutkevich 2021). Bidirectional models have this ability to read the whole sequence of words at once, thus each word in the sequence is processed at the same time. Through this contextual relations between words can be learned by the model, for example it can distinguish the word he in the sentence refers to Daniel, which is productive when analyzing sentiment in more than one sentences. Moreover,by utilizing BERT in NLP transfer learning (Brownlee 2017) we will be given this ability to fine-tune our model even with small amounts of data and get a convincible result.

Having said that, BERT is computationally expensive and has millions of hyper parameters (Yates et al.2021) [9]. Thus, the distilBERT model was used (Sanh 2019) [10]. This model is a distil version of the BERT with much less hyper parameters, and shorter fine-tuning time, medium computers are required as well. The custom dataset of classified English COVID19 answers was used or fine-tuning. The machine learning model was trained to classify the positive and the negative responses. For testing, we only use the textual data to see whether the model predict the sentiment of each sentence properly or not. The accuracy of our model on the test dataset was 80% which means the model can generalize the classification of the COVID19 responses in a good way.

The encoders and decoders are included in the structure of transformer as it is shown in (Fig 1.b). But the advantage of this kind of transformer is its principle on attention (Vaswani et al.2017) [11], which is:

$$\text{Attention } (Q, K, V) = \text{softmax}\left(\frac{QK^T}{\sqrt{d_k}}\right) = \text{VEc},$$

Where $Q$, $k$, and $V$ are query vector, key vector, and value vector respectively.

$$Q = X * W_q,$$
$$K = X * W_k,$$
$$V = X * W_v,$$

Where the output of embedding is $X$ and $W$s are the associated weights with each layer of the decoder's neural network which is a feed-forward one. Also, the multi-head attention can be modeled as various solo attention paralleled blocks with various workers.

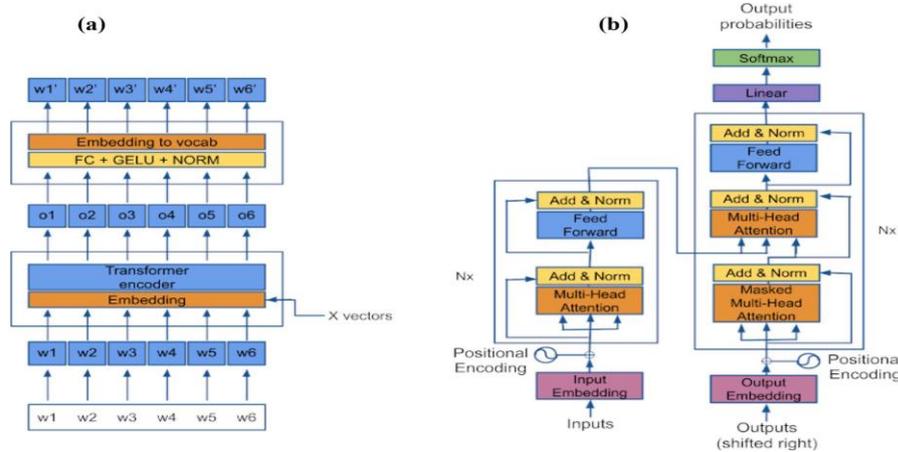

**Figure 1. Structure of Pretrained transformer model: a just encoder, b whole model (Tang et al. 2020) [12]**

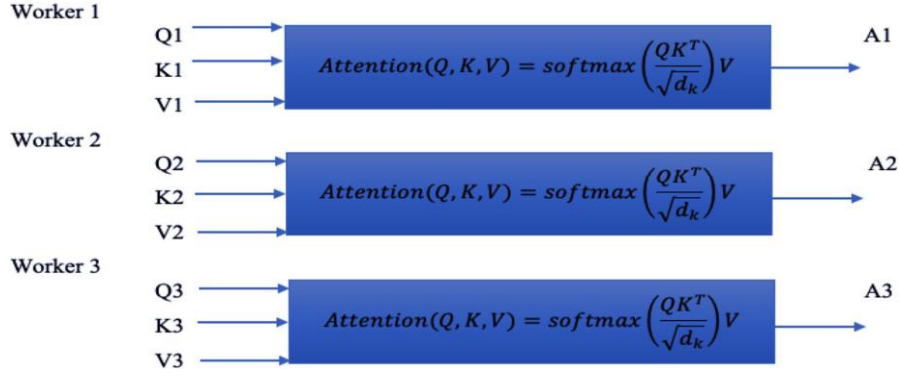

**Figure 2. Multi-attention block with parallel attention workers**

As Cen be seen in Figure 2. The positice point of using this technique is discovering the correlation among words. The aim of multi-head is to give us this ability to attend various part of the sequence every time, which means model can capture the positional information more productively, due to the fact that each head will attend various segment of the sequence. Furthermore, each head will capture different contextual information (Ho et al. 2019) [13].

So that, the model's fine tuning process (Sanh et al. 2019) [10] is to update and adjust weights of the neural network which is related to the distilBERT's attention layers with next context's information, which is desired for our study.

### 3-4. Metrics Measurement

The classification tasks and classifiers should be measured in the same way. The model of this study is a binary classifier, so to measure the performance common metrics such as Accuracy, Precision, Recall and $F1$ score (Wang, 2018) [14] were used.

$$Accuracy = \frac{TP+TN}{TP+TN+FP+FN} \quad (1)$$

$$Percision(i) = \frac{TP(i)}{TP(i)+FP(i)} \quad (2)$$

$$Percision(i) = \frac{TP(i)}{TP(i)+FN(i)} \quad (3)$$

$$F1\ score\ (i) = 2\frac{Percison(i)*Recal(i)}{Percison(i)+Recal(i)} \quad (4)$$

Where Accuracy gives the idea of common behavioural trend of the model, but does not consider the impact of generated bias from unbalanced dataset, so we use Precision as well, which is positive rate linked to false positives. Recall the positive rate linked to false negatives and F1 score which is second harmonic mean amalgamating the precision and recall metrics.

### 4. Results

The outcomes after applying distilBERT to our model is presented. After fine-tuning model to the mentioned dataset in section 2, the classifier pretrained with GLUE (General Language Understanding Evaluation benchmark) dataset and fine-tuned with three approaches (Liu et al.2020) [15].

The used hyper parameters for approach one are presented in Table 1. In this method, 80% of the dataset was used for training and 20% for testing, but in this way the bias was not taken into account because the positive and the negative are not balanced. Table 2 presents the confusion matrix of this approach. While the hyper parameters gained from the grid search method, there was the opportunity to get the best performance. From Table 2 it can be understood that sentences with negative sentiment are more prone to be classified well, while positive sentences have numerous wrong classification. Which means there is a necessity to apply a balanced subsets of dataset in order to get the best performance in classification.

**Table 1. Used hyperparameters for the first method**

| Hyperparameters | Value |
|---|---|
| Total number of training epochs | num_train_epoch = 7 |
| Batch size per device during training | Per_device_train_batch_size = 16 |
| Batch size for evaluation | Per_device_eval_batch_size = 64 |
| Number of warmup steps for learning rate | Warmup_steps = 500 |
| Strength of way decay | Weight_decay = 0.01 |

**Table 2. Confusion matrix for the first method**

| | Predict | | | positives/Negatives |
|---|---|---|---|---|
| | Class | Positives | Negatives | Total |
| Real | Positives | 69 | 56 | 125 |
| | Negatives | 17 | 187 | 204 |

In the second approach we used a balanced data set which contains 350 positives and 350 negatives. In Table 3 and 4 the used hyper parameters and confusion matrix can be seen respectively. As can be seen, the difference is just in the epochs' number. Table 5 presents that again negative sentences have high tendency to be classified correctly, while it should take into account that the positive sentences are better classified now compared to first approach.

**Table 3. Used hyperparameters for the second method**

| Hyperparameters | Value |
|---|---|
| Total number of training epochs | num_train_epoch = 9 |
| Batch size per device during training | Per_device_train_batch_size = 16 |
| Batch size for evaluation | Per_device_eval_batch_size = 64 |
| Number of warmup steps for learning rate | Warmup_steps = 500 |
| Strength of way decay | Weight_decay = 0.01 |

**Table 4. Confusion matrix for the second method**

| | Predict | | | positives/Negatives |
|---|---|---|---|---|
| | Class | Positives | Negatives | Total |
| Real | Positives | 84 | 41 | 125 |
| | Negatives | 18 | 186 | 204 |

In the third approach, the dataset was splatted to 90% for training and 10% for testing. In Table 5 and 6 the hyper parameters and confusion matrix for this approach can be seen in that order. Similar to the two pervious approaches the only thing which changed was epochs' number. As it is obvious from table 6 while the third approach had a superior performance, the data's number to test was reduced to 50%, and there is not much differences in performance in comparison to other approaches.

**Table 5. Used hyperparameters for the third method**

| Hyperparameters | Value |
|---|---|
| Total number of training epochs | num_train_epoch = 5 |
| Batch size per device during training | Per_device_train_batch_size = 16 |
| Batch size for evaluation | Per_device_eval_batch_size = 64 |
| Number of warmup steps for learning rate | Warmup_steps = 500 |
| Strength of way decay | Weight_decay = 0.01 |

**Table 6. Confusion matrix for the third method**

|  | Predict Class | Positives | Negatives | positives/Negatives Total |
|---|---|---|---|---|
| Real | Positives | 40 | 20 | 60 |
|  | Negatives | 8 | 90 | 98 |

Based on formulas (1), (2), (3) and (4) metrics for "negative sentiment" and "positive sentiment" can be calculated. Lastly, both are averaged to get an objective perspective of the model's performance. While the accuracy for classifying negative sentences is higher in the third approach, the second one with almost the same accuracy has a better precision and the first one has a good recall for the negative class and the poorest recall for the positive class, thus we can conclude that the first approach is not feasible to use.

**Table 7. Results obtained for "negative sentiment" class as the reference class**

| Approach | Accuracy | Precision | Recall | F1 score |
|---|---|---|---|---|
| First(data 80-20) | 0.777 | 0.769 | 0.915 | 0.835 |
| Second(data 50-50) | 0.822 | **0.823** | 0.905 | 0.860 |
| Third(data 90-10) | **0.824** | 0.817 | **0.918** | **0.863** |

**Table 8. Results obtained for "positive sentiment" class as the reference class**

| Approach | Accuracy | Precision | Recall | F1 score |
|---|---|---|---|---|
| First(data 80-20) | 0.777 | 0.799 | 0.548 | 0.658 |
| Second(data 50-50) | 0.822 | 0.818 | 0.655 | **0.749** |
| Third(data 90-10) | **0.824** | **0.832** | **0.678** | 0.746 |

The average for each of the metrics is calculated, the aim of doing is to examine the behavior of each model when facing unbalanced dataset, and from this we can see the approach 3 has the best performance in this situation. Also, it does worth to notice that the 3$^{rd}$ approach metrics' values are very close to that of the 2$^{nd}$ approach (Table 9). By considering the limitation in the number of used data, we came to conclusion that approach two is the best for classifying positive and negative sentiments in this study.

**Table 9. Average of the results gained for "negative sentiment" and "positive sentiment" classes**

| Approach | Accuracy | Precision | Recall | F1 score |
|---|---|---|---|---|
| First(data 80-20) | 0.777 | 0.784 | 0.731 | 0.746 |
| Second(data 50-50) | 0.822 | 0.820 | 0.780 | **0.804** |
| Third(data 90-10) | **0.824** | **0.824** | **0.798** | **0.804** |

## 5. Discussion

DistilBERT transformer is used in this study, while this transformer is simpler model of BERT transformer it has the same capacity as the bigger one. In a paper by Sanh et al. [10], Liu et al. [15], and Yu et al. [16], we understand that while distilBERT has lower computational complexity and lower parameters it presented a good performance. In this study, these features are worthwhile due to the fact that the size of our dataset is not much large and this lead us to prefer model with lower number of parameters.

Furthermore, we used Google API to translate the responses in Persian to English automatically. This step could be considered as a risk as the final classification to a great extend depends on the words in the sentence in the paper by Qiu et al. [17], they present how the automatic language translation is performed for a natural language processing model and how the utilization of encoder–decoder approach creates a desired sequence in the new desired code.

The Covid19 pandemic and its related lockdowns impacted the mental health of the individuals all around the world specifically students and staff and faculty's life and social health. Due to this fact some additional support is needed to be provided to tackle with the mental health issues which are the fruit of the COVID19 pandemic (Aqeel et al. 2021[18]; Cao et al. 2020 [19]; Abbas et al. 2019[20]).

In this study, an approach for analyzing open-text responses to some questions about the COVID19 and its impact is proposed. Since a face made decision is required in the case of pandemic, employing a number of experts for carrying out this task is not rational as it is expensive in terms of money and time. In the study, artificial intelligence is used which is taught from the experience of experts and then can detect the positive and negative cases in shorter time compared to human force.

## 6. Conclusion

In order to find the information about the sentiment in the text, sequence to sequence models can be applied. Due to the fact that the amount of data for training the model should be extremely large and the number of transformer's parameters are more than millions, one of the best approaches could be fine tuning, which can reduce the burden of training the model from scratch.

The main constraint of this study is that a pretrained model has been used as the base, and this condition could lead to bias based the used data in training the model from the very first steps. In the papers by Nadeem et al. (2004) [21], Sahlgren and Olsson (2019) [22], and Lovering et al. (2020) [23] describe how bias appear after fine tuning process in the language models. Nonetheless, it is obvious that it could be an alternative in cases the amount of data is limited and dealing with the cost of computation is a concern.

Evetually, the model is limited because the amount of data was limited. So we could not build our own embedding to code the input text to vector. For more sensitive study it is suggested to have the specific embedding for that task to get more accurate results, regarding the context of the text.

It is important to take into account that conducting a research by transformers has its own merits and demerits. With the aim of decreasing bias, the data collection is a must for future works and next, after collecting this data we can deploy them for training a customized embedding and train the distilBERT from the scratch as well.